\documentclass[letterpaper,11pt]{article}

\usepackage{amssymb}
\usepackage{amsmath}
\usepackage{amsthm}
\usepackage{bm}
\usepackage{color}
\usepackage[footnotesize]{caption}
\usepackage{graphicx}
\usepackage{algorithm}
\usepackage{algpseudocode}
\usepackage{subcaption}
\usepackage{tkz-graph}
\usetikzlibrary{calc}
\usepackage{stackengine}

\def\dashfill{\cleaders\hbox to .6em{-}\hfill}
\newcommand\dashline[1]{\hbox to #1{\dashfill\hfil}}

\usepackage{fancyhdr}
\fancypagestyle{plain}{
  \fancyhead{} 

  \chead{} 
  \cfoot{\thepage}
}

\usepackage{geometry}

\usepackage{epstopdf}

\begin{document}
\title{Parallel Complexity of Forward and Backward Propagation}
\author{Maxim Naumov \\ {\small NVIDIA, 2701 San Tomas Expressway, Santa Clara, CA 95050}}
\date{}
\maketitle

\begin{abstract}
We show that the forward and backward propagation can be formulated as a solution of lower and upper triangular systems of equations. For standard feedforward (FNNs) and recurrent neural networks (RNNs) the triangular systems are always block bi-diagonal, while for a general computation graph (directed acyclic graph) they can have a more complex triangular sparsity pattern. We discuss direct and iterative parallel algorithms that can be used for their solution and interpreted as different ways of performing model parallelism. Also, we show that for FNNs and RNNs with $k$ layers and $\tau$ time steps the backward propagation can be performed in parallel in O($\log k$) and O($\log k \log \tau$) steps, respectively. Finally, we outline the generalization of this technique using Jacobians that potentially allows us to handle arbitrary layers.   
\end{abstract}

\section{Introduction}

The forward and backward propagation are used in training of neural networks. The training is an optimization procedure that minimizes the loss function $\mathcal{L}$ over data samples $(\mathbf{x}^{*},\mathbf{z}^{*})$ in a data set $\mathcal{D}$ \cite{Bishop1995,Bishop2006,Goodfellow2016}. The loss function measures on average the error $\mathcal{E}(.,.)$ between the computed $\mathbf{z}^{(l)}$ and the correct solution $\mathbf{z}^{*}$, e.g. cross entropy error function
\begin{eqnarray}
\mathcal{E}(\mathbf{z}^{*},\mathbf{z}^{(l)}) &=& - \sum_{i=1}^{m} z_{i}^{*} \log \left( p_{i}  \right)  
\label{def_loss_component}
\end{eqnarray}
with the softmax function
\begin{eqnarray}
p_{i} = \frac{e^{z_{i}^{(l)}}}{\sum_{j=1}^{m} e^{z_{j}^{(l)}} }   
\label{def_loss_softmax}
\end{eqnarray}
and probability $\mathbf{p}=[p_{i}]$, target $\mathbf{z}^{*}=[z_{i}^{*}]$ as well as computed $\mathbf{z}^{(l)}=[z_{i}^{(l)}]$ at the output. 
\par
The forward propagation starts with an input $\mathbf{x}^{*}$. It applies an affine function $\theta_k$ followed by a component-wise application of a non-linear activation function $f_k$ to obtain an output of a layer. It proceeds sequentially through a composition of layers $k=1,...,l$ defining the neural network $\phi$. As a result it computes an output $\mathbf{z}^{(l)}$ at the final layer $l$, as shown below
\begin{equation}
\mathbf{z}^{(l)} = \phi(\mathbf{x}^{*}) =  \mathbf{f}_{l} \left\{  \theta_{l} \left(   \ldots \mathbf{f}_{2} \left[ \theta_{2} \left( \mathbf{f}_{1} \left(  \theta_{1} \left(  \mathbf{x}^{*}   \right) \right)      \right) \right]    \right) \right\}    
\label{def_forw_prop1}
\end{equation}
For instance, assuming that $\theta_{k} ( \mathbf{x} ) = W^{(k)} \mathbf{x} + \mathbf{b}^{(k)}$ we may write 
\begin{equation}
\mathbf{z}^{(l)} = \mathbf{f}_{l} \left\{  W^{(l)}  \ldots \mathbf{f}_{2} \left[ W^{(2)} \mathbf{f}_{1} (  W^{(1)} \mathbf{x}^{*} + \mathbf{b}^{(1)} ) + \mathbf{b}^{(2)}  \right] \ldots + \mathbf{b}^{(l)} \right\} 
\label{def_forw_prop2}
\end{equation}
where matrix of weights $W^{(k)} \in \mathbb{R}^{m \times n}$ and vector of bias $\mathbf{b}^{(k)} \in \mathbb{R}^{m}$, with dimensions $m$ and $n$ being consistent but potentially different across layers.
\par
The backward propagation starts with an error $\bm{\epsilon}^{T}=\nabla \mathcal{E}_{l}^{T} = [\partial \mathcal{E}/ \partial z_1^{(l)},...,\partial \mathcal{E}/ \partial z_m^{(l)}]$ at the final layer. It uses a chain rule 
\begin{eqnarray}
\nabla \mathcal{E}_{k-1}
&=& J_{\theta_k}^{T} \left( \nabla \mathcal{E}_{k} \circ \mathbf{f}_k' \right)  
\label{gradient_z_hidden_layer} 
\end{eqnarray}
to find the corresponding error at the previous level \cite{Hinton1986,Rumelhart1986,Werbos1990}, where $J_{\theta_k} = [\partial y_i^{(k)} / \partial z_j^{(k-1)}]$ is a Jacobian matrix, $\mathbf{f}_k' = [f'(y_1^{(k)}),....,f'(y_m^{(k)})]^{T}$ and $\mathbf{y}^{(k)} = \theta_k (\mathbf{z}^{(k-1)})$, while $\circ$ denotes the Hadamard (component-wise) product. Notice that $\mathbf{z}^{(k-1)}$ denotes both the output of $(k-1)$-th and the input to $k$-th layer, with $\mathbf{z}^{(0)}=\mathbf{x}^{*}$. The backward propagation proceeds sequentially backwards through a composition of layers $k=l,...,1$ defining the neural network $\phi$. As a result it computes errors $\mathbf{v}^{(k)} = \nabla \mathcal{E}_{k} \circ \mathbf{f}_k'$ at all layers, as shown below 
\begin{equation}
\mathbf{v}^{(0)} = J_{\phi}^{T} \bm{\epsilon} =  J_{\theta_1}^{T} \{ \ldots  \mathbf{f}'_{l-2} \circ  ( J_{\theta_{l-1}}^{T} [\mathbf{f}'_{l-1} \circ ( J_{\theta_l}^{T} \mathbf{v}^{(l)} ) ] ) \}
\label{def_back_prop1}
\end{equation}
where $J_{\phi} = [\partial \phi_i / \partial \epsilon_j]$ is a Jacobian of the neural network \cite{Naumann2012,Pearlmutter2008,Spivak1971}. 
\par
For instance, for the cross entropy error function in \eqref{def_loss_component} we can write
\begin{equation}
\nabla \mathcal{E}_{l} = \mathbf{p} - \mathbf{z}^{*}
\label{gradient_z_output_layer}
\end{equation}
and for function $\theta_k$ in \eqref{def_forw_prop2} we can write
\begin{equation}
\mathbf{v}^{(0)} = W^{(1)^{T}} \left\{ \ldots \mathbf{f}_{l-2}' \circ \left( W^{(l-1)^{T}} \left[ \mathbf{f}_{l-1}' \circ (  W^{(l)^{T}} \mathbf{v}^{(l)} )  \right] \right) \right\} 
\label{def_back_prop2}
\end{equation}
\par
The errors $\mathbf{v}^{(k)}$ can then be used to update coefficients of functions $\theta_k$. In particular, in \eqref{def_forw_prop2} these coefficients are the weights $W^{(k)}$ and bias $\mathbf{b}^{(k)}$, where we can write
\begin{eqnarray}
\Delta W^{(k)} = \frac{\partial \mathcal{E}}{\partial w^{(k)}}
&=& \left( \nabla \mathcal{E}_{k} \circ \mathbf{f}_k' \right) \left( \frac{\partial \mathbf{y}^{(k)}}{\partial w^{(k)}} \right) \label{gradient_w_hidden_layer} \\
\Delta \mathbf{b}^{(k)} = \frac{\partial \mathcal{E}}{\partial \mathbf{b}^{(k)}}
&=& \left( \nabla \mathcal{E}_{k} \circ \mathbf{f}_k' \right) \label{gradient_b_hidden_layer} 
\end{eqnarray}
\par
We point out that the last term in \eqref{gradient_w_hidden_layer} can have different interpretations. Let us consider deriving $\Delta W^{(k)}$ for \eqref{def_forw_prop2} using component-wise derivatives and a composition of Jacobians. In the former case, the last term will be a vector and the final result will be computed as an outer product. In the latter case, it will be a three dimensional tensor that will collapse to a matrix under multiplication. In both cases, the final result will be the same, though. 
\par
Finally, notice that in practice the data samples are often organized into mini-batches to take a better advantage of parallel computing platforms \cite{Pascal2017,Volta2017}. A smaller mini-batch can often achieve higher test accuracy, while a larger mini-batch can attain higher performance \cite{Das2016,Devarakonda2017,Goyal2017,Keskar2016,You2017}. Therefore, the stochastic gradient descent and its many variants \cite{Bottou2016}, often average the updates across the mini-batch, as shown below 
\begin{eqnarray}
W_{new}^{(k)}   &=& W^{(k)} - \frac{\alpha}{r} \Delta \bar{W}^{(k)} \\
\Delta \bar{W}^{(k)} &=& \sum_{i=1}^{r} \Delta W^{(k,i)}
\end{eqnarray}
where $r$ is the mini-batch size and $\alpha$ is the learning rate \cite{Schmidhuber2015}.
\par
Notice that the data samples inside a mini-batch are independent and can be processed in parallel, which is often referred to as \textit{data parallelism}. On the other hand, the forward and backward propagation traversals through the layers are in principle sequential. Therefore, taking advantage of \textit{model parallelism}, parallelism across layers, is more difficult. 
\par
In this paper we will show that in fact the forward and backward propagation can be interpreted as a solution of block bi-diagonal triangular systems of equations. This transformation allows us to use an entire class of well developed parallel numerical techniques to solve these systems. Further, it allows us to choose whether we want to solve these problems approximately or exactly. In this novel approach, the many different parallel schemes developed for this problem in linear algebra can now be seen as simply different ways of performing \textit{model parallelism} in neural networks. 

\section{Block Bi-Diagonal Triangular Systems}
Let a neural network $\phi$ be defined by a composition of affine $\theta_k : \mathbb{R}^{n} \rightarrow \mathbb{R}^{m}$ and non-linear $f_k : \mathbb{R} \rightarrow \mathbb{R}$ functions that are applied component-wise on the output, for $k=1,...,l$ layers. Further, let the affine function be a sum $\theta_k = \lambda_k + \mathbf{b}$ of linear function $\lambda_k : \mathbb{R}^{n} \rightarrow \mathbb{R}^{m}$ and a vector of constants $\mathbf{b} \in \mathbb{R}^m$. Also, recall that $\mathbf{y}^{(k)} = \theta_k (\mathbf{z}^{(k-1)})$ where $\mathbf{z}^{(k-1)}$ is the input to $k$-th layer, with $\mathbf{z}^{(0)}=\mathbf{x}^{*}$, and note that Jacobian $J_{\theta_k} = J_{\lambda_k}$. 
\par
Notice that we can write forward propagation \eqref{def_forw_prop1} as the solution of the block bi-diagonal lower triangular system $\bar{L}\bar{\mathbf{z}}=\bar{\mathbf{b}}$, written in extended form below
\begin{equation}
\left(
\begin{array}{rrrrr}
D^{(0)}        \\
-\lambda^{(1)} &   D^{(1)} \\
         &  -\lambda^{(2)} & D^{(2)} \\
         &           & \ldots  & \ldots \\
         &           &                    & -\lambda^{(l)}          & D^{(l)} \\
\end{array}
\right)
\left(
\begin{array}{c}
\mathbf{z}^{(0)}   \\
\mathbf{z}^{(1)} \\
\mathbf{z}^{(2)} \\
\ldots \\
\mathbf{z}^{(l)}
\end{array}
\right)
=
\left(
\begin{array}{c}
\mathbf{x}^{*} \\
\mathbf{b}^{(1)} \\
\mathbf{b}^{(2)} \\
\ldots \\
\mathbf{b}^{(l)}
\end{array}
\right)
\label{forw_prop_sys_jacobian}
\end{equation}
where $D^{(0)}=I$ and diagonal matrices $D^{(k)^{-1}}=\text{diag}(\mathbf{f}_{k}) \text{diag}(\mathbf{y}_{k})^{-1}$ for $k=1,...,l$. 
\par
Also, notice that we can write backward propagation \eqref{def_back_prop1} as the solution of the block bi-diagonal upper triangular system $\bar{R}\bar{\mathbf{v}}=\bar{\mathbf{c}}$, written in extended form below
\begin{equation}
\left(
\begin{array}{rr@{}rrr@{}r@{}}
E^{(0)\phantom{^{T}}}       & -J_{\lambda_1}^{T} \\
        & E^{(1)\phantom{^{T}}}     & -J_{\lambda_2}^{T}\\
        &             & \ldots & \ldots \\
        &             &        & E^{(l-1)} & -J_{\lambda_l}^{T} \\
        &             &        &           & E^{(l)\phantom{^{T}}}     \\
\end{array}
\right)
\left(
\begin{array}{@{}c@{}}
\mathbf{v}^{(0)} \\
\mathbf{v}^{(1)} \\
\ldots           \\
\mathbf{v}^{(l-1)} \\
\mathbf{v}^{(l)}
\end{array}
\right)
=
\left(
\begin{array}{c}
\mathbf{0} \\
\mathbf{0} \\
\ldots \\
\mathbf{0} \\
\bm{\epsilon}
\end{array}
\right)
\label{back_prop_sys_jacobian}
\end{equation}
where $\bm{\epsilon} = \nabla \mathcal{E}_{l}$, $E^{(0)}=I$ and diagonal matrices $E^{(k)^{-1}}=\text{diag}(\mathbf{f}'_{k})$ for $k=1,...,l$. 
\par
Notice that it might appear that \eqref{forw_prop_sys_jacobian} and \eqref{back_prop_sys_jacobian} define two linear systems, which is contradictory to the non-linear nature of the forward and backward propagation process in neural networks. However, notice that in reality the non-linearity is implicitly hidden in the diagonal elements $D^{(k)}$ and $E^{(k)}$ of these systems. In fact, we can represent it in this form only because the non-linear activation function is applied component-wise, and therefore \textit{if we know the point $y_{i}^{(k)} \neq 0$, the function $f(y_{i}^{(k)})$ and its derivative $f'(y_{i}^{(k)})$ evaluations, then we can always find a scalars $\alpha = f(y_{i}^{(k)})/y_{i}^{(k)}$ and $\beta = f'(y_{i}^{(k)})$, such that the results can be obtained using a scaling with those constants}. 
\par
Notice that in forward propagation the points $\mathbf{y}^{(k)}$ and $\mathbf{f}_k$ are not known ahead of time. However, they can potentially be estimated from a previous pass over the data. This approximation can be effective in the later stages of training, when we are in the region close to the solution. Therefore, we can approximate forward propagation by solving \eqref{forw_prop_sys_jacobian}.
\par
Notice that in backward propagation the starting points $\mathbf{y}^{(k)}$ are in fact known and $\mathbf{f}'(\mathbf{y}^{(k)})$ can be pre-computed ahead of time. Therefore, we can perform backward propagation exactly by solving \eqref{back_prop_sys_jacobian}.   
\par
We point out that the theory we have developed handles arbitrary functions $\theta_k$ as well as most popular activation functions $f_k$. In particular, notice that when solving triangular systems \eqref{forw_prop_sys_jacobian} with forward and \eqref{back_prop_sys_jacobian} with backward substitution, we always multiply with $f(y)$ or $f'(y)$ diagonal terms, respectively. 
Therefore, the reformulation of backward propagation in \eqref{back_prop_sys_jacobian} works independent of whether $f'(y) \neq 0$ or $f'(y)=0$. The reformulation of forward propagation in \eqref{forw_prop_sys_jacobian} works with points $y \neq 0$ and $f(y)=0$ for analogous reasons. Also, notice that when $y=0$ and $f(y)=0$ the desired effect can be achieved by simply setting diagonal element to unity. However, the reformulation of forward propagation in \eqref{forw_prop_sys_jacobian} does break down when $y=0$ and $f(y) \neq 0$, because there is no constant that multiplied by zero produces a non-zero result. When using floating point numbers we are often close, but rarely exactly at the numerical zero, therefore the approach is likely to succeed  
in practice. Nonetheless, it is clear that very careful and thoughtful algorithm design is required to handle all the corner cases correctly.
\par
Also, notice that a similar formulation can be written for an arbitrary compute graph, that can be expressed as a directed acyclic graph. However, in general the triangular systems will no longer be block bi-diagonal and can have a more complex sparsity pattern.
\par
There is a number of algorithms in numerical linear algebra that have been developed to solve block bi-diagonal triangular systems. In general we can split them into direct methods that solve the system exactly and iterative methods that solve the system approximately. All of these numerical schemes can now be seen as simply different ways of performing \textit{model parallelism} in neural networks. Also, notice that working with multiple data samples organized into mini-batches is equivalent to finding a solution of a set of shifted systems
\begin{eqnarray}
(\bar{L} + \bar{D}_i)\bar{\mathbf{z}}=\bar{\mathbf{b}}_i \label{forw_prop_sys_shifted} \\
(\bar{R} + \bar{E}_i)\bar{\mathbf{v}}=\bar{\mathbf{c}}_i \label{back_prop_sys_shifted}
\end{eqnarray}
for $i=1,...,r$, which is a well studied problem and allows for additional \textit{data parallelism}. 

\subsection{FNNs}

Let us assume that we are given a data sample $(\mathbf{x}^{*},\mathbf{z}^{*})$ and that a feedforward neural network $\phi$ is defined by an affine function $\theta_{k} ( \mathbf{x} ) = W^{(k)} \mathbf{x} + \mathbf{b}^{(k)}$ and an activation function $f_k$ for $k=1,...,l$ layers.
\par
Let $\bm{\epsilon} = \nabla \mathcal{E}_{l}$ be the error gradient at the output layer $l$. Recall that the error $\mathbf{v}^{(l)} = \bm{\epsilon} \circ \mathbf{f}_l'$ can be propagated through the neural network by repeated applications of the formula 
\begin{eqnarray}
\mathbf{v}^{(k-1)} &=& ( W^{(k)^{T}} \mathbf{v}^{(k)} )   \circ \mathbf{f}' (\mathbf{y}^{(k-1)}) \phantom {111111} 
\end{eqnarray}
for levels $k=l,...,2$ \cite{Naumov2017}. 
\par
Notice that following \eqref{forw_prop_sys_jacobian} and \eqref{back_prop_sys_jacobian} we can write forward propagation as the solution of the block bi-diagonal lower triangular system $\bar{L}\bar{\mathbf{z}}=\bar{\mathbf{b}}$, written in extended form below
\begin{equation}
\left(
\begin{array}{rrrrr}
D^{(0)}        \\
-W^{(1)} &   D^{(1)} \\
         &  -W^{(2)} & D^{(2)} \\
         &           & \ldots  & \ldots \\
         &           &                    & -W^{(l)}          & D^{(l)} \\
\end{array}
\right)
\left(
\begin{array}{c}
\mathbf{z}^{(0)}   \\
\mathbf{z}^{(1)} \\
\mathbf{z}^{(2)} \\
\ldots \\
\mathbf{z}^{(l)}
\end{array}
\right)
=
\left(
\begin{array}{c}
\mathbf{x}^{*} \\
\mathbf{b}^{(1)} \\
\mathbf{b}^{(2)} \\
\ldots \\
\mathbf{b}^{(l)}
\end{array}
\right)
\label{forw_prop_sys}
\end{equation}
while backward propagation as the solution of the block bi-diagonal upper triangular system $\bar{R}\bar{\mathbf{v}}=\bar{\mathbf{c}}$, written in extended form below
\begin{equation}
\left(
\begin{array}{rr@{}rrr@{}r@{}}
E^{(0)\phantom{^{T}}}       & -W^{(1)^{T}} \\
        & E^{(1)\phantom{^{T}}}     & -W^{(2)^{T}}\\
        &             & \ldots & \ldots \\
        &             &        & E^{(l-1)} & -W^{(l)^{T}} \\
        &             &        &           & E^{(l)\phantom{^{T}}}     \\
\end{array}
\right)
\left(
\begin{array}{@{}c@{}}
\mathbf{v}^{(0)} \\
\mathbf{v}^{(1)} \\
\ldots           \\
\mathbf{v}^{(l-1)} \\
\mathbf{v}^{(l)}
\end{array}
\right)
=
\left(
\begin{array}{c}
\mathbf{0} \\
\mathbf{0} \\
\ldots \\
\mathbf{0} \\
\bm{\epsilon}
\end{array}
\right)
\label{back_prop_sys}
\end{equation}
where $D^{(0)}=E^{(0)}=I$ and diagonal matrices $D^{(k)}=[d_{ii}^{(k)}]$ and $E^{(k)}=[e_{ii}^{(k)}]$ have elements
\begin{eqnarray}
d_{ii}^{(k)} &=& \frac{y_{i}^{(k)}}{f(y_{i}^{(k)})} \text{ and } \label{forw_diag}\\
e_{ii}^{(k)} &=& \frac{1}{f'(y_{i}^{(k)})}, \label{back_diag}  
\end{eqnarray}
respectively, for $k=1,...,l$.

\subsection{RNNs}

Let us assume that we are given a sequence $(\mathbf{x}^{(*,s)},\mathbf{z}^{(*,s)})$ as a data sample and that a recurrent neural network $\phi$ is defined by an affine function $\theta_{k} ( \mathbf{x}^{(k,s)} ) = W^{(k)}\mathbf{x}^{(k-1,s)}+U^{(k)}\mathbf{x}^{(k,s-1)}+\mathbf{b}^{(k)}$ and an activation function $f_k$ for $k=1,...,l$ layers and $s=1,...,\tau$ time steps.
\par
Let $\bm{\epsilon}^{(s)} = = \nabla \mathcal{E}_{l,s}$ be the error gradient at the output layer $l$ at time step $s$. Recall that the error $\mathbf{v}^{(l,s)} = \bm{\epsilon}^{(s)} \circ \mathbf{f}_{l,s}'$ can be propagated through the neural network by repeated applications of the formula 
\begin{eqnarray}
\mathbf{v}^{(k-1,s)} &=&   \left( W^{(k)^{T}} \mathbf{v}^{(k,s)} + U^{(k-1)^{T}} \mathbf{v}^{(k-1,s+1)} \right) \circ \mathbf{f}' (\mathbf{y}^{(k-1,s)}) 
\end{eqnarray}
for levels $k=l,...,2$ and time steps $s=\tau,...,1$, where we consider the terms for time $s+1>\tau$ to be zero \cite{Naumov2017}. 
\par
Notice that following \eqref{forw_prop_sys_jacobian} and \eqref{back_prop_sys_jacobian} we can write forward propagation as the solution of the block bi-diagonal lower triangular system $\tilde{L}\tilde{\mathbf{z}}=\tilde{\mathbf{b}}$, written in extended form below
\begin{equation}
\left(
\begin{array}{rrrrrr}
\bar{L}^{(1)}  \\
-\bar{U}^{(2)} &  \bar{L}^{(2)}  \\
               & -\bar{U}^{(3)} & \bar{L}^{(3)}  \\
               &                & \ldots        & \ldots \\
               &                &               & -\bar{U}^{(\tau)} & \bar{L}^{(\tau)}  \\
\end{array}
\right)
\left(
\begin{array}{c}
\bar{\mathbf{z}}^{(1)} \\
\bar{\mathbf{z}}^{(2)} \\
\bar{\mathbf{z}}^{(3)} \\
\ldots \\
\bar{\mathbf{z}}^{(\tau)}
\end{array}
\right)
=
\left(
\begin{array}{c}
\bar{\mathbf{b}}^{(1)} \\
\bar{\mathbf{b}}^{(2)} \\
\bar{\mathbf{b}}^{(3)} \\
\ldots \\
\bar{\mathbf{b}}^{(\tau)}
\end{array}
\right)
\label{forw_prop_sys_rnn}
\end{equation}
while backward propagation as the solution of the block bi-diagonal upper triangular system $\tilde{R}\tilde{\mathbf{v}}=\tilde{\mathbf{c}}$, written in extended form below
\begin{equation}
\left(
\begin{array}{@{}r@{}r@{}r@{}rr@{}}
\bar{R}^{(1)\phantom{^{T}}} & -\bar{U}^{(2)^{T}} \\
              & \bar{R}^{(2)\phantom{^{T}}}      & -\bar{U}^{(3)^{T}} \\
              &                    & \ldots            & \ldots \\
              &                    &                   & \bar{R}^{(\tau-1)} & -\bar{U}^{(\tau)^{T}} \\
              &                    &                   &                 & \bar{R}^{(\tau)\phantom{^{T}}}     \\
\end{array}
\right)
\left(
\begin{array}{@{}c@{}}
\bar{\mathbf{v}}^{(1)} \\
\bar{\mathbf{v}}^{(2)} \\
\ldots \\
\bar{\mathbf{v}}^{(\tau-1)} \\
\bar{\mathbf{v}}^{(\tau)}
\end{array}
\right)
=
\left(
\begin{array}{c}
\bar{\mathbf{c}}^{(1)} \\
\bar{\mathbf{c}}^{(2)} \\
\ldots \\
\bar{\mathbf{c}}^{(\tau-1)} \\
\bar{\mathbf{c}}^{(\tau)}
\end{array}
\right)
\label{back_prop_sys_rnn}
\end{equation}
where 
\begin{equation}
\bar{L}^{(s)} =
\left(
\begin{array}{rrrrr}
\phantom{-}D^{(0,s)}        \\
-W^{(1)} &   D^{(1,s)} \\
         &  -W^{(2)} & D^{(2,s)} \\
         &           & \ldots  & \ldots \\
         &           &                    & -W^{(l)}          & D^{(l,s)} \\
\end{array}
\right)
\text{, }
\bar{\mathbf{z}}^{(s)} =
\left(
\begin{array}{@{}c@{}}
\mathbf{z}^{(0,s)}   \\
\mathbf{z}^{(1,s)} \\
\mathbf{z}^{(2,s)} \\
\ldots \\
\mathbf{z}^{(l,s)}
\end{array}
\right)
\text{ and }
\bar{\mathbf{b}}^{(s)} = 
\left(
\begin{array}{@{}c@{}}
\mathbf{x}^{(*,s)} \\
\mathbf{b}^{(1)} \\
\mathbf{b}^{(2)} \\
\ldots \\
\mathbf{b}^{(l)}
\end{array}
\right)
\label{forw_prop_sub_sys_rnn}
\end{equation}

\begin{equation}
\bar{R}^{(s)} =
\left(
\begin{array}{@{}rrr@{}rr@{}r}
E^{(0,s)}       & -W^{(1)^{T}} \\
        & E^{(1,s)}     & -W^{(2)^{T}}\\
        &             & \ldots & \ldots \\
        &             &        & E^{(l-1,s)} & -W^{(l)^{T}} \\
        &             &        &           & E^{(l,s)}     \\
\end{array}
\right)
\text{, }
\bar{\mathbf{v}}^{(s)} =
\left(
\begin{array}{@{}c@{}}
\mathbf{v}^{(0,s)} \\
\mathbf{v}^{(1,s)} \\
\ldots \\
\mathbf{v}^{(l-1,s)} \\
\mathbf{v}^{(l,s)}
\end{array}
\right)
\text{ and }
\bar{\mathbf{c}}^{(s)} =
\left(
\begin{array}{@{}c@{}}
\mathbf{0} \\
\mathbf{0} \\
\ldots \\
\mathbf{0} \\
\bm{\epsilon}^{(s)}
\end{array}
\right)
\label{back_prop_sub_sys_rnn}
\end{equation}
with $D^{(0,s)}=E^{(0,s)}=I$ and diagonal matrices $D^{(k,s)}=[d_{ii}^{(k,s)}]$ and $E^{(k,s)}=[e_{ii}^{(k,s)}]$ so that
\begin{eqnarray}
d_{ii}^{(k,s)} &=& \frac{y_{i}^{(k,s)}}{f(y_{i}^{(k,s)})} \text{ and } \label{forw_diag_rnn}\\
e_{ii}^{(k,s)} &=& \frac{1}{f'(y_{i}^{(k,s)})}, \label{back_diag_rnn}  
\end{eqnarray}
respectively, for $k=2,...,l$. Finally, letting $U^{(0)}=0$ we may write
\begin{equation}
\bar{U}^{(s)} =
\left(
\begin{array}{rrrrr@{}r}
U^{(0)} \\
        & U^{(1)} \\
        &         & U^{(2)} \\
        &         &         & \ldots \\
        &         &         &        & U^{(l)} \\
\end{array}
\right)
\end{equation}
that in fact does not depend on time steps $s$.
\par
Notice that the systems \eqref{forw_prop_sys_rnn} and \eqref{back_prop_sys_rnn} are block bi-diagonal, with their diagonal blocks $\bar{L}^{(s)}$ and $\bar{R}^{(s)}$ given in  \eqref{forw_prop_sub_sys_rnn} and \eqref{back_prop_sub_sys_rnn}, respectively. Further, notice that $\bar{L}^{(s)}$ and $\bar{R}^{(s)}$ are themselves block bi-diagonal systems. In fact, they correspond to \eqref{forw_prop_sys} and \eqref{back_prop_sys} for a particular time step $s$. 

\section{Parallel Algorithms} 
There is a number of algorithms in numerical linear algebra that have been developed to solve block bi-diagonal triangular systems. In general we can split them into direct methods that solve the system exactly and iterative methods that solve the system approximately. All of these numerical schemes can now be seen as simply different ways of performing \textit{model parallelism} in neural networks. Also, notice that working with multiple data samples organized into mini-batches can simply be seen as performing the solution with shifted systems \eqref{forw_prop_sys_shifted} and \eqref{back_prop_sys_shifted}, which allows for additional \textit{data parallelism}.

\subsection{Direct Methods}

The direct methods include (parallel) cyclic reduction \cite{Arbenz1998,Buzbee1970,Hockney1965,Hockney1988}, nested dissection \cite{Dongarra1987,Johnsson1985}, spike-based \cite{Dongarra1984,Polizzi2006,Polizzi2007,Sameh1978}, domain decomposition-based \cite{Lou1989,Naumov2009,Naumov2010} and many other schemes \cite{Alvarado1993,Egecioglu1989,Sameh1977,Stone1973,Zhang2010}.  
\par
For example, let us illustrate the parallel solution of the system \eqref{forw_prop_sys} and \eqref{back_prop_sys} using a variant of cyclic reduction. For clarity let us assume that we have a neural network with an odd number of levels $l > 1$. The handling of the case with even number of layers is similar. 
\par
Notice that using scaling $S$ and permutation $P$ matrices \footnote{Please refer to appendix for the exact expression.}, we may write an equivalent system $(S\bar{L}P)(P^T\bar{\mathbf{z}})=S\bar{\mathbf{b}}$ to \eqref{forw_prop_sys} in block bi-diagonal lower triangular form as
\begin{equation}
\left(
\begin{array}{rrrr|rrrr}
D^{(1)}  &           &          & \\
-B^{(3)} &   D^{(3)} &          & \\
         &   \ldots  & \ldots   & \\
         &           & -B^{(l)} & D^{(l)} \\
\hline
         &           &          &         & D^{(0)}  &         &            & \\
         &           &          &         & -B^{(2)} & D^{(2)} &            & \\
         &           &          &         &          & \ldots  & \ldots     & \\
         &           &          &         &          &         & -B^{(l-1)} & D^{(l-1)} \\  
\end{array}
\right)
\left(
\begin{array}{l}
\mathbf{z}^{(1)} \\
\mathbf{z}^{(3)} \\
\ldots \\
\mathbf{z}^{(l)} \\
\hline
\mathbf{x}^{*} \\
\mathbf{z}^{(2)} \\
\ldots \\
\mathbf{z}^{(l-1)}
\end{array}
\right)
=
\left(
\begin{array}{l}
\mathbf{g}^{(1)} \\
\mathbf{g}^{(3)} \\
\ldots \\
\mathbf{g}^{(l)} \\
\hline
\mathbf{x}^{*} \\
\mathbf{g}^{(2)} \\
\ldots \\
\mathbf{g}^{(l-1)}
\end{array}
\right)
\label{forw_prop_sys_reordered}
\end{equation}
where off-diagonal blocks $B^{(k)} = W^{(k)}D^{(k-1)^{-1}}W^{(k-1)}$, right-hand-side $\mathbf{g}^{(0)} = \mathbf{x}^{*}$, $\mathbf{g}^{(1)} = W^{(1)}\mathbf{x}^{*}+\mathbf{b}^{(1)}$ and $\mathbf{g}^{(k)} = W^{(k)}D^{(k-1)^{-1}}\mathbf{b}^{(k-1)}+\mathbf{b}^{(k)}$ for $k=2,...,l$. 
\par
Also, that using scaling $T$ and permutation $Q$ matrices$^1$, we may write an equivalent system $(T\bar{R}Q)(Q^T\bar{\mathbf{v}})=T\bar{\mathbf{c}}$ to \eqref{back_prop_sys} in block bi-diagonal upper triangular form as
\begin{equation}
\left(
\begin{array}{@{}r@{}r@{}rr|r@{}r@{}rr@{}}
E^{(0)}\phantom{^{T}}  & -C^{(2)^{T}}         &               & \\
         &    \ldots & \ldots   &               & \\
         &           & E^{(l-3)}\phantom{^{T}}& -C^{(l-1)^{T}}& \\
         &           &          &  E^{(l-1)}\phantom{^{T}}    & \\
\hline
         &           &          &               & E^{(1)}\phantom{^{T}} & -C^{(3)^{T}} &            & \\
         &           &          &               &         & \ldots       & \ldots     & \\
         &           &          &               &         &              & E^{(l-2)}\phantom{^{T}}  & -C^{(l)^{T}} \\
         &           &          &               &         &              &            &  E^{(l)}\phantom{^{T}}    
\end{array}
\right)
\left(
\begin{array}{c@{}}
\mathbf{v}^{(0)} \\
\ldots \\
\mathbf{v}^{(l-3)} \\
\mathbf{v}^{(l-1)} \\
\hline
\mathbf{v}^{(1)} \\
\ldots \\
\mathbf{v}^{(l-2)} \\
\mathbf{v}^{(l)}
\end{array}
\right)
=
\left(
\begin{array}{c@{}}
\mathbf{0} \\
\ldots \\
\mathbf{0} \\
\mathbf{h}^{(l-1)} \\
\hline
\mathbf{0} \\
\ldots \\
\mathbf{0} \\
\mathbf{h}^{(l)} \\
\end{array}
\right)
\label{back_prop_sys_reordered}
\end{equation}
where off-diagonal blocks $C^{(k)^{T}} = W^{(k-1)^{T}}E^{(k-1)^{-1}}W^{(k)^{T}}$, right-hand-side $\mathbf{h}^{(l)} = \bm{\epsilon}$, $\mathbf{h}^{(l-1)} = W^{(l)^{T}}E^{(l)^{-1}}\bm{\epsilon}$ and $\mathbf{h}^{(k-2)} = \mathbf{c}^{(k-2)}+W^{(k-1)^{T}}E^{(k-1)^{-1}}\mathbf{c}^{(k-1)}=\mathbf{0}$ for $k=2,...,l$. 
\par
Notice that both \eqref{forw_prop_sys_reordered} and \eqref{back_prop_sys_reordered} decouple into two independent smaller systems of half the size of the original. Since we can apply this approach recursively, we can perform approximate forward and exact backward propagation in parallel in O($\log l$) steps. 
\par
Notice that based on equivalence of original linear system in \eqref{forw_prop_sys} and \eqref{forw_prop_sys_reordered}, we can state that forward propagation in \eqref{def_forw_prop2} is equivalent to two independent forward propagations
\begin{eqnarray}
\mathbf{z}^{(l)} &=& \mathbf{f}_{l} \left\{  B^{(l)}  \ldots \mathbf{f}_{5} \left[ B^{(5)} \mathbf{f}_{3} (  B^{(3)} \mathbf{g}^{(1)} + \mathbf{g}^{(3)} ) + \mathbf{g}^{(5)}  \right] + \mathbf{g}^{(l)} \right\} 
\label{def_forw_prop2_odd} \\
\mathbf{z}^{(l-1)} &=& \mathbf{f}_{l-1} \left\{  B^{(l-1)}  \ldots \mathbf{f}_{4} \left[ B^{(4)} \mathbf{f}_{2} (  B^{(2)} \mathbf{g}^{(0)} + \mathbf{g}^{(2)} ) + \mathbf{g}^{(4)}  \right] + \mathbf{g}^{(l-1)} \right\} 
\label{def_forw_prop2_even}
\end{eqnarray}
with modified weights $B^{(k)}$ for $k=2,...,l$ and inputs $\mathbf{g}^{(0)}$ and $\mathbf{g}^{(1)}$ at the initial layer.
\par
Also, based on equivalence of original linear system in \eqref{back_prop_sys} and \eqref{back_prop_sys_reordered}, we can state that backward propagation in \eqref{def_back_prop2} is equivalent to two independent backward propagations
\begin{eqnarray}
\mathbf{v}^{(0)} &=& C^{(2)^{T}} \left\{ \ldots \mathbf{f}_{l-5}' \circ \left( C^{(l-3)^{T}} \left[ \mathbf{f}_{l-3}' \circ (  C^{(l-1)^{T}} \mathbf{v}^{(l-1)} )  \right] \right) \right\} 
\label{def_back_prop2_even} \\
\mathbf{v}^{(1)} &=& C^{(3)^{T}} \left\{ \ldots \mathbf{f}_{l-4}' \circ \left( C^{(l-2)^{T}} \left[ \mathbf{f}_{l-2}' \circ (  C^{(l)^{T}} \mathbf{v}^{(l)} )  \right] \right) \right\} 
\label{def_back_prop2_odd}
\end{eqnarray}
with modified weights $C^{(k)}$ for $k=2,...,l$ and errors $\mathbf{h}^{(l)}$ and $\mathbf{h}^{(l-1)}$ at the final layer. 
\par
The solution of systems \eqref{forw_prop_sys_rnn} and \eqref{back_prop_sys_rnn} is very similar, with the exception that each diagonal block $\bar{L}^{(s)}$ in \eqref{forw_prop_sub_sys_rnn} and $\bar{R}^{(s)}$ in \eqref{back_prop_sub_sys_rnn} is itself a block bi-diagonal system. Therefore, the systems \eqref{forw_prop_sys_rnn} and \eqref{back_prop_sys_rnn} can be solved in parallel in O($(\log k)(\log \tau)$) steps. 

\subsection{Iterative Methods}

The iterative methods include fixed-point schemes, such as Jacobi, Gauss-Seidel and Richardson iterations, as well as Krylov subspace methods, such as CG, BiCGStab and GMRES, among many others algorithms \cite{Barret1994,Saad2003}. Notice that in this particular application we have non-symmetric systems, where either the strictly upper or lower triangular part is zero. Also, the matrices are often very large and it might be prohibitively expensive to store the vector subspace generated throughout iterations. Therefore Jacobi, Richardson and BiCGStab methods are natural choices for solving these systems.
\par
\par
For example, let us illustrate the parallel solution of the systems \eqref{forw_prop_sys} and \eqref{back_prop_sys} using the Jacobi iteration. Let the splitting $\bar{L}=\bar{D}-\bar{B}$ and $\bar{R}=\bar{E}-\bar{C}$ be such that $\bar{D}$ and $\bar{E}$ are the block-diagonal, while $\bar{B}$ and $\bar{C}$ are the strictly block lower and upper triangular parts of the original matrix, respectively. Then, we can write Jacobi iteration for \eqref{forw_prop_sys} as
\begin{equation}
\mathbf{z}_{i+1} = \bar{D}^{-1}\bar{B} \mathbf{z}_{i} + \bar{D}^{-1}\mathbf{b}
\end{equation}
and Jacobi-like iteration for \eqref{back_prop_sys} as
\begin{equation}
\mathbf{v}_{i+1} = \bar{E}^{-1}\bar{C} \mathbf{v}_{i} + \bar{E}^{-1}\mathbf{c}
\end{equation}
Notice that the sufficient condition for this iteration to converge is that matrices $\bar{L}$ and $\bar{R}$ are diagonally dominant, which can be checked ahead of time. This computation is in fact embarrassingly parallel, in the sense that every row of matrix $\bar{D}^{-1}\bar{B}$ and $\bar{E}^{-1}\bar{C}$ is independent of others. However, multiple iterations might be required to achieve the desired accuracy.
\par
Let us also briefly touch on the parallel solution of the systems \eqref{forw_prop_sys} and \eqref{back_prop_sys} using BiCGStab method. The convergence of BiCGStab is governed by the distribution of eigenvalues of the coefficient matrix \cite{Greenbaum1997}. It is well known that the eigenvalues of a triangular matrix are simply the elements on its diagonal \cite{Horn1999}. Therefore, it is natural to scale the coefficient matrix by the diagonal, so that all eigenvalues are clustered around one. This can be achieved by rewriting the linear systems \eqref{forw_prop_sys} as
\begin{equation}
(\bar{D}^{-1}\bar{L})\bar{\mathbf{z}}=\bar{D}^{-1}\bar{\mathbf{b}}
\label{bicgstab_scaled_forw_sys}
\end{equation}
and \eqref{back_prop_sys} as
\begin{equation}
(\bar{E}^{-1}\bar{R})\bar{\mathbf{v}}=\bar{E}^{-1}\bar{\mathbf{c}}
\label{bicgstab_scaled_back_sys}
\end{equation}
The scaling by the block diagonal can be seen as preconditioning of the BiCGStab method. In fact more general preconditioning techniques can further speed up convergence, but might be prohibitively expensive in terms of computation and storage per iteration.
\par
Notice that when we apply BiCGStab to solve the linear systems \eqref{bicgstab_scaled_forw_sys} and \eqref{bicgstab_scaled_back_sys} the most compute intensive step in the algorithm is once again a matrix-vector multiplication of the form $\bar{\mathbf{q}}=(\bar{D}^{-1}\bar{L})\bar{\mathbf{p}}$ and $\bar{\mathbf{q}}=(\bar{E}^{-1}\bar{R})\bar{\mathbf{p}}$, respectively. The former can be written in extended form as
\begin{equation}
\left(
\begin{array}{c}
\mathbf{q}^{(0)} \\
\mathbf{q}^{(1)} \\
\mathbf{q}^{(2)} \\
\ldots \\
\mathbf{q}^{(l)}
\end{array}
\right)
=
\left(
\begin{array}{rrrrr}
I        \\
D^{(1)^{-1}}W^{(1)} &   I \\
         &  D^{(2)^{-1}}W^{(2)} & I \\
         &           & \phantom{..........}\ldots  & \ldots \\
         &           &                    & D^{(l)^{-1}}W^{(l)}          & I \\
\end{array}
\right)
\left(
\begin{array}{c}
\mathbf{p}^{(0)} \\
\mathbf{p}^{(1)} \\
\mathbf{p}^{(2)} \\
\ldots \\
\mathbf{p}^{(l)}
\end{array}
\right)
\label{forw_prop_matvec}
\end{equation}
while the latter can be written in extended form as
\begin{equation}
\left(
\begin{array}{c}
\mathbf{q}^{(0)} \\
\mathbf{q}^{(1)} \\
\ldots \\
\mathbf{q}^{(2)} \\
\mathbf{q}^{(l)}
\end{array}
\right)
=
\left(
\begin{array}{rrrrrr}
I\phantom{^{T}}       & E^{(0)^{-1}}W^{(1)^{T}} \\
        &  I\phantom{^{T}}     & E^{(1)^{-1}}W^{(2)^{T}}\\
        &             & \ldots & \ldots \\
        &             &        & I & E^{(l-1)^{-1}}W^{(l)^{T}} \\
        &             &        &           & I\phantom{^{T}}     \\
\end{array}
\right)
\left(
\begin{array}{c}
\mathbf{p}^{(0)} \\
\mathbf{p}^{(1)} \\
\ldots \\
\mathbf{p}^{(2)} \\
\mathbf{p}^{(l)}
\end{array}
\right)
\label{back_prop_matvec}
\end{equation}
Notice that if we distribute this matrix and the corresponding vectors by rows across multiple nodes then only pair-wise communication between nodes is required to compute the result in parallel. However, multiple iterations of BiCGStab are required for convergence. 

\subsection{Mini-Batch and Shifted Systems}

Finally, notice that in neural networks we are often working with a mini-batch of data, and therefore we would need to solve a set of shifted systems, such as the ones shown in  \eqref{forw_prop_sys_shifted} and \eqref{back_prop_sys_shifted}, to extract additional \textit{data parallelism}. Notice that the only difference between these systems are the diagonal blocks $D^{(k)}$ and $E^{(k)}$, which are based on a particular data sample, while off diagonal blocks $B^{(k)}$ and $C^{(k)}$ remain the same. Therefore, the generalization of our approach to a set of shifted systems is feasible for direct methods. In the case of iterative methods, we can address it by relying on the slightly modified implementations of the block Krylov subspace methods  \cite{Guennouni2003,Naumov2012,OLeary1980,Simoncini1996}.

\subsection{Hybrid Methods}

Notice that we might choose to combine the direct and iterative methods to solve a single linear system. This could be especially useful for larger systems resulting from recurrent networks, such as \eqref{forw_prop_sys_rnn} and \eqref{back_prop_sys_rnn}. For instance, we can choose to solve diagonal blocks $\bar{L}^{(s)}$ in \eqref{forw_prop_sub_sys_rnn} and $\bar{R}^{(s)}$ in \eqref{back_prop_sub_sys_rnn} directly, while the entire system is solved iteratively. In the future, we plan to experiment with all of these strategies.

\section{Time, Work and Memory Complexity}
Let us now make a more detailed analysis of the complexity of parallel forward and backward propagation algorithm. For instance, let us illustrate it for the fully connected neural network with weight matrices $W^{(k)} \in \mathbb{R}^{n_{k+1} \times n_{k}}$, where forward and backward propagation is defined in \eqref{def_forw_prop2} and \eqref{def_back_prop2}, respectively. For clarity let us assume that we have a neural network with an odd number of levels $l > 1$. The handling of the case with even number of layers is similar. Also, we will assume a standard theoretical PRAM (CREW) model for the formal analysis \cite{JaJa1992}.
\par
The sequential complexity of forward and backward propagation can be written as
\begin{eqnarray}
\sum_{k=1}^{l} (n_{k+1} n_{k} + \gamma n_{k+1}) r  \label{seqcomplexity} 
\end{eqnarray}
where $r$ is the batch size and $\gamma$ is some scalar constant that measures the amount of operations required to evaluate the non-linear activation function. Notice that the first term in \eqref{seqcomplexity} corresponds to the matrix-matrix multiplication, e.g. evaluation of the affine function $\theta$ or $J_{\theta}^{T}$, while the last term corresponds to the evaluation of the component-wise non-linear activation function $\mathbf{f}$.
\par
Notice that if we assume that $n_{k} \approx n$ for $k=1,...,l$ then we can state that the sequential complexity is O$(ln^2r)$.

\subsection{Direct Methods}
  
The work complexity of parallel forward propagation in \eqref{def_forw_prop2_odd} - \eqref{def_forw_prop2_even} as well as backward propagation in \eqref{def_back_prop2_even} - \eqref{def_back_prop2_odd} can be written as 
\begin{eqnarray}
\sum_{k=2}^{l} ( n_{k+1} n_{k} n_{k-1} + n_{k+1} n_{k} )
+ \sum_{k=1}^{(l-1)/2} (n_{i+1} n_{i-1} + \gamma n_{i+1} ) r
+ \sum_{k=1}^{(l-1)/2} (n_{i+2} n_{i}   + \gamma n_{i+2} ) r 
\label{parworkcomplexity} 
\end{eqnarray}
where index $i=2k$. The first term in \eqref{parworkcomplexity} corresponds to the overhead required to construct the intermediate weights $B^{(k)}$ and right-hand-sides $\mathbf{g}^{(k)}$ as well as weights $C^{(k)}$ and right-hand-sides $\mathbf{h}^{(k)}$ for forward as well as backward propagation, respectively. The last terms in \eqref{parworkcomplexity} correspond to two independent propagations, with the newly constructed weights.
\par
Therefore, the time complexity of forward and backward propagation is
\begin{eqnarray}
\max_k ( n_{k+1} n_{k} n_{k-1} + n_{k+1} n_{k} )
+ \max_{j \in \{i,i+1\}} \sum_{k=1}^{(l-1)/2} (n_{j+1} n_{j-1} + \gamma n_{j+1} ) r
\label{partimecomplexity}
\end{eqnarray}
where $p=l$ processors are available for computation.  
\par
Notice that if we assume that $n_{k} \approx n$ for $k=1,...,l$ then the parallel work and time complexity is O$(ln^3 + ln^2r)$ and O$(n^3 + \frac{l}{2}n^2r)$, respectively. 
\par
Further, applying this approach recursively, we can write for time complexity
\begin{eqnarray}
\underbrace{n^3 + \ldots n^3}_{\log l} + n^2 r
\label{recursive_complexity}
\end{eqnarray}
and therefore obtain that the parallel work and time complexity can be expressed as  O$(ln^3\log l + ln^2r)$ and O$(n^3\log l + n^2r)$, respectively. Notice that the first term in \eqref{recursive_complexity} corresponds to the computation of new weights and right-hand-sides in parallel for $\log l$ levels of recursion, while the last term corresponds to $l$ independent propagations across a single layer that can be computed in parallel at the leafs of the recursion.
\par
Finally, recall that the algorithm constructs the intermediate weights $B^{(k)}$ and right-hand-sides $\mathbf{g}^{(k)}$ as well as weights $C^{(k)}$ and right-hand-sides $\mathbf{h}^{(k)}$ for forward as well as backward propagation, respectively. Notice that if the algorithm is applied recursively it can overwrite these intermediate weights, with the new weights at a deeper level of the recursion, in other words, there is no need to preserve the weights across multiple levels of the recursion. Therefore, we may conclude that the parallel forward and backward propagation require no more than double the amount of memory of the sequential forward and backward propagation, respectively.  
\par
It is important to point out that based on the above analysis the parallel propagation algorithm presented above becomes practical when the mini-batch size $r \ge n$ or $l/\log l \ge n$, otherwise the overhead of computing intermediate weights might be prohibitively expensive. The parallel algorithm might also be relevant when the weight matrices are sparse and the complexity of sparse matrix-vector and matrix-matrix multiplication is more closely associated with number of non-zeros, rather than the dimensions of the matrix. 

\subsection{Iterative Methods}
The work complexity of approximating forward and backward propagation in \eqref{def_forw_prop2} and \eqref{def_back_prop2} using an iterative method defined by the matrix-vector multiplications in \eqref{forw_prop_matvec} and \eqref{back_prop_matvec}, respectively, can be written as 
\begin{eqnarray}
\left[ \sum_{k=1}^{l} (n_{k+1} n_{k} + \gamma n_{k+1}) r \right] \eta  \label{parworkcomplexity_it} 
\end{eqnarray}
where $\eta$ is the number of iterations taken to convergence by an iterative method. Notice that similar to the sequential case in \eqref{seqcomplexity} the first term in \eqref{parworkcomplexity_it} corresponds to the matrix-matrix multiplication, e.g. evaluation of the affine function $\theta$ or $J_{\theta}^{T}$, while the last term corresponds to the diagonal scaling associated with the activation function $\mathbf{f}$.
\par
However, notice that all rows in matrix-vector multiplication \eqref{forw_prop_matvec} and \eqref{back_prop_matvec} can be processed independently. Therefore, the time complexity of forward and backward propagation can be expressed as
\begin{eqnarray}
\left[ \max_k ( n_{k+1} n_{k} + \gamma n_{k+1} ) r \right] \eta
\label{partimecomplexity_it}
\end{eqnarray}
where $p=l$ processors are available for computation.  
\par
Notice that if we assume that $n_{k} \approx n$ for $k=1,...,l$ then the parallel work and time complexity is O$(ln^2r\eta)$ and O$(n^2r\eta)$, respectively. 
\par
Finally, notice that the iterative methods, such as Jacobi, Richardson or BiCGStab, only require extra storage for a few intermediate vectors during iterations. Let us say that in the worst case $2 \le \delta \le 7$ such vectors are in fact needed (the exact parameter depends on the iterative method of choice). We may conclude that the parallel forward and backward propagation require no more than $\delta (l n r )$ the amount of memory of the sequential propagation. Notice that the memory required for the latter corresponds to the term in parenthesis.  
\par
It is important to point out that based on the above analysis the approximate parallel propagation algorithm presented above becomes practical when the number of iterations required for convergence $\eta < l$. 
 
\section{Conclusion} 
In this paper we have developed a framework for representing forward and backward propagation as solution of triangular systems. This representation has allowed us to develop novel parallel schemes for forward and backward propagation. In particular, we have illustrated how to apply a variation of cyclic reduction to parallelize the propagation process. We have analysed its parallel time, work and memory complexity and outlined the criteria of when the algorithm is practically useful. 
\par
We highlight that the approach presented in this paper is not limited to any particular algorithm. In fact, it outlines a general framework that can be used to design parallel propagation schemes based on different numerical linear algebra techniques. In particular, many of the direct and iterative methods used to solve triangular systems can now be interpreted as different ways of performing model parallelism.    

\section{Acknowledgements}
The author would like to acknowledge Cris Cecka, Aditya Devarakonda, Joe Eaton, Michael Garland, Boris Ginsburg, Oleksii Kuchaiev and Nathan Whitehead for their insightful comments and suggestions.

\newpage 

\section{Appendix}
\par
Finally, for completeness we state the scaling and permutation matrices used to obtain \eqref{forw_prop_sys_reordered} and \eqref{back_prop_sys_reordered} from \eqref{forw_prop_sys} and \eqref{back_prop_sys}, respectively. The scaling matrix $S$ is defined as
\begin{equation}
S=
\left(
\begin{array}{cccccccc}
W^{(1)}D^{(0)^{-1}}  & I \\
         &   & W^{(3)}D^{(2)^{-1}} &   I \\
         &   &                     &   & \ldots  & \ldots \\
         &   &                     &   &         &        & W^{(l)} D^{(l-1)^{-1}}   & I\\
I        \\
0        & W^{(2)}D^{(1)^{-1}}&   I \\
         &                    &     &  \ldots  & \ldots \\
         &                    &     &          &        & W^{(l)} D^{(l-1)^{-1}}   & I & 0\\
\end{array}
\right)
\nonumber
\end{equation}
and the scaling matrix $T$ is defined as 
\begin{equation}
T=
\left(
\begin{array}{ccc@{}c@{}cccc@{}}
I & W^{(1)^{T}}E^{(1)^{-1}} &  \\
  &                         & I & W^{(3)^{T}}E^{(3)^{-1}} &  \\
  &                         &   &                         &  \ldots & \ldots \\
  &                         &   &                         &         &        & I & W^{(l)^{T}} E^{(l)^{-1}}  \\
0 & I                       & W^{(2)^{T}}E^{(2)^{-1}} \\
  &                         &                         &  \ldots  & \ldots \\
  &                         &                         &          &        & I & W^{(l-1)^{T}}E^{(l-1)^{-1}} \\  
  &                         &                         &          &        &   &                             & I \\ 
\end{array}
\right)
\nonumber
\end{equation}
while permutation matrices that rearranges odd and even variables are defined as
\begin{equation}
P^{T}=
\left(
\begin{array}{cccccccccc}
0 & I \\
  &   & 0 & I \\
  &   &   &   &  \ldots & \ldots \\
  &   &   &   &         &        & 0 & I \\
I & 0 \\
  &   & I & 0 \\
  &   &   &   &  \ldots & \ldots \\ 
  &   &   &   &         &        & I & 0 \\  
\end{array}  
\right)
\text{ and }
Q^{T}=
\left(
\begin{array}{cccccccccc}
I & 0 \\
  &   & I & 0 \\
  &   &   &   &  \ldots & \ldots \\ 
  &   &   &   &         &        & I & 0 \\
0 & I \\
  &   & 0 & I \\
  &   &   &   &  \ldots & \ldots \\
  &   &   &   &         &        & 0 & I \\  
\end{array}  
\right)
\nonumber
\end{equation}


\begin{thebibliography}{00}

\bibitem{Alvarado1993} {\sc F. L. Alvarado and R. Schreiber}, {\em Optimal Parallel Solution of Sparse Triangular Systems}, SIAM J. Scientific Computing, Vol. 14, pp. 446-460, 1993.

\bibitem{Arbenz1998} {\sc P. Arbenz and M. Hegland}, {\em The Stable Parallel Solution of General Narrow Banded Linear Systems}, High Performance Algorithms for Structured Matrix Problems, pp. 47-73, 1998.

\bibitem{Barret1994} {\sc R. Barrett, M. Berry, T. F. Chan, J. Demmel, J. Donato, J. Dongarra, V. Eijkhout, R. Pozo, C. Romine, H. van der Vorst}, {\em Templates for the Solution of Linear Systems: Building Blocks for Iterative Methods}, SIAM, Philadelphia, PA, 1994.

\bibitem{Bishop1995} {\sc C. Bishop}, {\em Neural Networks for Pattern Recognition}, Oxford University Press, January, 1995.

\bibitem{Bishop2006} {\sc C. Bishop}, {\em Pattern Recognition and Machine Learning}, Springer, January, 2006. 

\bibitem{Bottou2016} {\sc L. Bottou, F. E. Curtis and J. Nocedal}, {\em Optimization Methods for Large Scale Machine Learning}, Technical Report, arXiv:1606.04838, 2016.

\bibitem{Buzbee1970} {\sc B. L. Buzbee, G. H. Golub, and C. W. Nielson}, {\em On Direct Methods for Solving Poisson's Equation}, SIAM Journal on Numerical Analysis, Vol. 7, pp. 627-656, 1970.

\bibitem{Das2016} {\sc D. Das, S. Avancha, D. Mudigere, K. Vaidynathan, S. Sridharan, D. Kalamkar, B. Kaul and P. Dubey}, {\em Distributed Deep Learning Using Synchronous Stochastic Gradient Descent}, Technical Report, arXiv:1602.06709, 2016.

\bibitem{Devarakonda2017} {\sc A. Devarakonda, M. Naumov and M. Garland}, {\em AdaBatch: Adaptive Batch Sizes for Training Deep Neural Networks}, Technical Report, arXiv:1712.02029, 2017.

\bibitem{Dongarra1984} {\sc J. J. Dongarra and A. H. Sameh}, {\em On Some Parallel Banded System Solvers}, Parallel Computing, Vol. 1, pp. 223-235, 1984.

\bibitem{Dongarra1987} {\sc J. J. Dongarra and S. L. Johnsson}, {\em Solving Banded Systems on a Parallel Processor} Parallel Computing, Vol. 5, pp. 219-246, 1987.

\bibitem{Egecioglu1989} {\sc O Egecioglu, C. K. Koc and A. J. Laub}, {\em A Recursive Doubling Algorithm for Solution of Tridiagonal Systems on Hypercube Multiprocessors}, Vol. 27, pp. 95-108, 1989.

\bibitem{George1973} {\sc J. A. George}, {\em  Nested Dissection of a Regular Finite Element Mesh}, SIAM Journal on Numerical Analysis, Vol. 10, pp. 345–363, 1973. 

\bibitem{Guennouni2003} {\sc A. El Guennouni, K. Jbilou and H. Sadok}, {\em A Block Version of BiCGStab for Linear Systems with Multiple RHS}, ETNA, Vol. 16, pp. 129-142, 2003.

\bibitem{Goodfellow2016} {\sc I. Goodfellow, Y. Bengio and A. Courville}, {\em Deep Learning}, MIT Press, 2016.

\bibitem{Goyal2017} {\sc P. Goyal, P. Dollar, R. B. Girshick, P. Noordhuis, L. Wesolowski, A. Kyrola, A. Tulloch, Y. Jia and K. He}, {\em Accurate, Large Minibatch SGD: Training ImageNet in 1 Hour}, Technical Report, arXiv:1706.02677, 2017.

\bibitem{Greenbaum1997} {\sc  A. Greenbaum}, {\em Iterative Methods for Solving Linear Systems}, Frontiers in Applied Mathematics, SIAM, Philadelphia, PA, 1997.

\bibitem{Hinton1986} {\sc D. E. Rumelhart, G. E. Hinton and R. J. Williams}, {\em Learning Internal Representations by Error Propagation}, Parallel Distributed Processing: Explorations in the Microstructure of Cognition, Foundations, Vol. 1, MIT Press, Cambridge, MA.

\bibitem{Hockney1965} {\sc	R. W. Hockney}, {\em A Fast Direct Solution of Poisson Equation Using Fourier Analysis}, Journal of the ACM, Vol. 12, pp. 95-113, 1965.

\bibitem{Hockney1988} {\sc R. W. Hockney and C. R. Jesshope}, {\em Parallel Computers 2: Architecture, Programming and Algorithms}, Institute of Physics Publishing, 1988.

\bibitem{Horn1999} {\sc R. A. Horn and  C. R. Johnson}, {\em Matrix Analysis}, Cambridge University Press, New York, NY, 1999.

\bibitem{JaJa1992} {\sc J. JaJa}, {\em An Introduction to Parallel Algorithms}, Addison-Wesley, 1992.

\bibitem{Johnsson1985} {\sc S. L. Johnsson}, {\em Solving Narrow Banded Systems on Ensemble Architectures}, Journal ACM Transactions on Mathematical Software, Vol. 11, pp. 271-288, 1985.

\bibitem{Keskar2016} {\sc N. S. Keskar, D. Mudigere, J. Nocedal, M. Smelyanskiy and P. T. P. Tang}, {\em On Large-Batch Training for Deep Learning: Generalization Gap and Sharp Minima}, technical Report, arXiv:1609.04836, 2016.

\bibitem{Lou1989} {\sc G. Lou}, {\em Parallel Methods for Solving Linear Systems via Overlapping Decomposition}, M.S. Thesis, University of Illinois, Urbana Champagne, 1989.

\bibitem{Naumann2012} {\sc U. Naumann}, {\em The Art of Differentiating Computer Programs: An Introduction to Algorithmic Differentiation}, SIAM, Philadelphia, PA, 2012.

\bibitem{Naumov2009} {\sc M. Naumov and A. Sameh}, {\em A Tearing-based Hybrid Parallel Banded Linear System Solver}, Journal of Computational and Applied Mathematics, Vol. 226, 306-318, 2009.

\bibitem{Naumov2010} {\sc M. Naumov, M. Manguoglu and A. Sameh}, {\em A Tearing-based Hybrid Parallel Sparse Linear System Solver}, Journal of Computational and Applied Mathematics, Vol. 234, 3025-3038, 2010.

\bibitem{Naumov2012} {\sc M. Naumov}, {\em Preconditioned Block-Iterative Methods on GPUs}, Proc. 83rd Annual Meeting of the International Association of Applied Mathematics and Mechanics, PAMM, Vol. 12, 11-14, 2012. 

\bibitem{Naumov2017} {\sc M. Naumov}, {\em Feedforward and Recurrent Neural Networks Backward Propagation and Hessian in Matrix Form}, Technical Report, arXiv:1709.06080 [cs.LG], 2017.

\bibitem{OLeary1980} {\sc D. P. O'Leary}, {\em The Block Conjugate Gradient Algorithm and Related Methods}, Linear Algebra Appl., Vol. 29, pp. 293–322, 1980.

\bibitem{Pearlmutter2008} {\sc B. Pearlmutter and J. Siskind}, {\em Reverse Mode AD in a Functional Framework: Lambda the Ultimate Backpropagator}, ACM TOPLAS, 2008.

\bibitem{Polizzi2006} {\sc E. Polizzi and A. H. Sameh}, {\em A Parallel Hybrid Banded System Solver: The SPIKE Algorithm}, Parallel Computing, Vol. 32, pp. 177-194, 2006.

\bibitem{Polizzi2007} {\sc E. Polizzi and A. H. Sameh}, {\em SPIKE: A Parallel Environment for Solving Banded Linear Systems}, Comput. Fluids, Vol. 36, pp. 113-120, 2007.

\bibitem{Rumelhart1986} {\sc D. E. Rumelhart, G. E. Hinton and R. J. Williams}, {\em Learning Representations by Back-Propagating Errors}, Nature, Vol. 323, pp. 533-536, 1986.

\bibitem{Saad2003} {\sc Y. Saad}, {\em Iterative Methods for Sparse Linear Systems}, SIAM, Philadelphia, PA, 2nd Ed., 2003.

\bibitem{Sameh1977} {\sc A. H. Sameh and R. P. Brent}, {\em Solving Triangular Systems on a Parallel Computer}, SIAM J. Numerical Analysis, Vol. 14, pp. 1101-1113, 1977.

\bibitem{Sameh1978} {\sc A. H. Sameh and D. J. Kuck}, {\em On Stable Parallel Linear System Solvers}, Journal of the ACM, Vol. 25, pp. 81-91, 1978.   

\bibitem{Schmidhuber2015} {\sc J. Schmidhuber}, {\em Deep Learning in Neural Networks: An Overview}, Neural Networks, Vol. 61, pp. 85-117, 2015. 

\bibitem{Simoncini1996} {\sc V. Simoncini and E. Gallopoulos}, {\em Convergence Properties of Block GMRES and Matrix Polynomials}, Linear Algebra and its Applications, Vol. 247, pp. 97-119, 1996.

\bibitem{Spivak1971} {\sc M. Spivak}, {\em Calculus On Manifolds: A Modern Approach To Classical Theorems Of Advanced Calculus}, Westview Press, 5th edition, 1971.

\bibitem{Stone1973} {\sc H. S. Stone}, {\em An Efficient Parallel Algorithm for the Solution of a Tridiagonal Linear System of Equations}, Journal of the ACM, Vol. 20, pp. 27-38, 1973. 

\bibitem{Werbos1990} {\sc P. Werbos}, {\em BackPropagation Through Time: What it Does and How to Do It?}, Proc. IEEE, 1990. 

\bibitem{You2017} {\sc Y. You, Igor. Gitman and B. Ginsburg}, {\em Scaling SGD Batch Size to 32K for ImageNet Training}, Technical Report, arXiv:1708.03888, 2017.

\bibitem{Zhang2010} {\sc Y. Zhang, J. Cohen and J. D. Owens}, {\em Fast Tridiagonal Solvers on the GPU}, Proc. 15th ACM SIGPLAN Symposium on Principles and Practice of Parallel Programming, pp. 127-136, 2010. 

\bibitem{Pascal2017} {\sc Nvidia}, {\em Nvidia Tesla P100 GPU}, Pascal Architecture White Paper, 2016.

\bibitem{Volta2017} {\sc Nvidia}, {\em Nvidia Tesla V100 GPU}, Volta Architecture White Paper, 2017.

\end{thebibliography}
\end{document}